# Coordinate System Selection for Minimum Error Rate Training in Statistical Machine Translation

**Lijiang Chen**
School of Chinese Language and Culture
Nanjing Normal Uuiversity
Nanjing, 210097, China
`ljchen97@126.com`

**Abstract**

Minimum error rate training (MERT) is a widely used training procedure for statistical machine translation. A general problem of this approach is that the search space is easy to converge to a local optimum and the acquired weight set is not in accord with the real distribution of feature functions. This paper introduces coordinate system selection (RSS) into the search algorithm for MERT. Contrary to previous approaches in which every dimension only corresponds to one independent feature function, we create several coordinate systems by moving one of the dimensions to a new direction. The basic idea is quite simple but critical that the training procedure of MERT should be based on a coordinate system formed by search directions but not directly on feature functions. Experiments show that by selecting coordinate systems with tuning set results, better results can be obtained without any other language knowledge.

## 1 Introduction

Statistical machine translation technologies convert a natural language into another natural language automatically by using large-scale corpus-based statistical models. From noise channel model (Brown, 1990), maximum entropy model (Och, 2002) to minimum error rate training model (Och, 2003), statistical machine translation systems continuously revise decision-making methods for obtaining translations, increasing the system performance gradually. Meanwhile, the machine translation units start to transit from words (Brown, 1990) to phrases (Koehn, 2003). For example, being the most widely used machine translation system, Moses contains two main technologies which are minimum error rate training model and statistical phrase-based translation (Koehn, 2007).

Minimum error rate training is a method proposed by Och (2003) to obtain all weights of feature functions according to the translation error rate. Unlike log-linear model based on maximum entropy, the MERT model does not use language model, distortion model and sentence length et al. directly to evaluate translations, but evaluates translations by a standard evaluation method (such as BLEU). Experiments show that when using the same evaluation method to carry out training and evaluating, the evaluation results can get better scores (Och, 2003). Although this method is not a real optimum algorithm, it can adapt translations to different evaluation methods. So it is more beneficial than the traditional method to obtain higher scores.

Assuming the translation $e$ of the source language sentence $f$ has a reference sentence $r$. The function $E(r,e)$ indicate the number of errors received comparing $r$ with $e$. Multiple errors of sentences can be cumulative:

$$E(r_1^s, e_1^s) = \sum_{s=1}^{S} E(r_s, e_s) \quad (1)$$

The goal of minimum error rate training is to optimize the parameters, so as to obtain the smallest number of errors on the development set $f_1^S$ with given reference translations $\hat{e}_1^S$ and a set of N-best candidate translations $C_s = \{e_{s,1}, ..., e_{s,K}\}$ for each sentence $f_s$.

$$\hat{\lambda}_1^I = \arg\min_{\lambda_i^I} \left\{ \sum_{s=1}^{S} E(r_s, \hat{e}(f_s; \lambda_1^M)) \right\} \quad (2)$$

$$= \arg\min_{\lambda_i^I} \left\{ \sum_{s=1}^{S} \sum_{k=1}^{K} E(r_s, e_{s,k}) \delta(\hat{e}(f_s; \lambda_1^M), e_{s,k}) \right\}$$

with

$$\hat{e}(f_s; \lambda_1^M) = \arg\max_{e \in C_d} \left\{ \sum_{m=1}^{M} \lambda_m h_m(e | f_s) \right\} \quad (3)$$

A standard algorithm for optimizing the parameters of eq. 3 is Koehn's Coordinate Descent (KCD). When training parameters, greedy algorithm is used to adjust the parameters of a particular dimension by turn, while other dimensions of the parameters are constant.

Although KCD can achieve a global optimum on one dimension each time, it can not guarantee that the entire feature set converge to a global optimum in all dimensions (Moore, 2008). In this paper, we introduce coordinate system selection (RSS) into KCD, so that we can change search directions in order to get better weight set to reflect the real distribution of feature functions. For closed test results and open test results[1] are often consistent, closed test results are used to choose better coordinate systems (RS) for the open test.

The goal of this paper is to investigate the role of coordinate system selection to optimize the parameters for MERT. Section 2 introduces Koehn's Coordinate Descent for minimum error training. Section 3 elaborates RSS-MERT including the theoretical basis of RSS and search progress. Section 4 gives the specific experimental results of three kinds of RSS. Section 5 presents the related research in recent years about improving MERT. Section 6 concludes the experiment, and gives future research directions of RSS.

## 2 Koehn's Coordinate Descent (KCD)

KCD is a variant of coordinate descent algorithm that, at each iteration, moves along the coordinate which allows for the most progress in the objective (Cer, 2008). Liu (2008) gives the progress steps of KCD (See Algorithm 1).

## 3 Coordinate systems Selection Model

### 3.1 KCD Independence Assumption

In KCD algorithm the search space is formed by a number of dimensions that correspond to all the feature functions. It tries to find a better scoring point in the parameter space by optimizing one parameter along one direction while keeping other parameters fixed (och, 2003).

---

[1] We use the tuning set to do the closed test and the test set to do the open test.

| | |
|---|---|
| Algotithm 1 | Koehn's Coordinate Descent method to find the minimum error rate |

1  iter=1;lasterror=0;newerror=1;epsilon=0.001
2  While (iter<Maxiter and abs (lasterror-newerror) > epsilon)
3      for d=1 to M // for each dimension
4       for s=1 to S // for each sentence in training corpus
5         Compute critical value and delta error
6       endfor
7     merge all critical value and sort them
8     compute error within each pair of non-identical boundaries
9     select the weight $\lambda$ as the midpoint of the interval corresponding to the lowest error
10    end for
11    iter++
12    lasterror=newerror
13    newerror=lowest error
14 end while

Given that all dimensions form a coordinate system, each dimension of the coordinate system corresponds to one and only one feature function. As each dimension of the coordinate system in the search process is independent, one to one correspondence coordinate system assumes that every two feature functions are independent, namely makes the assumption of independence.

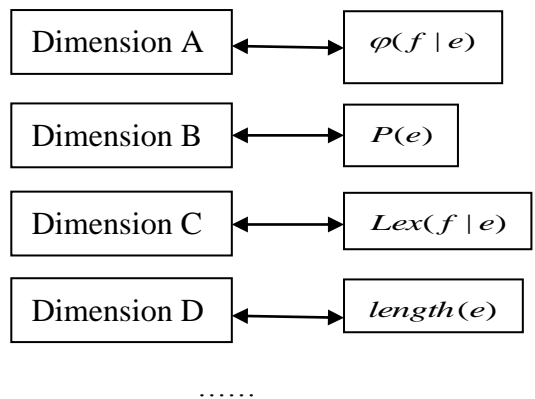

……

However, not all features functions are independent. Some feature functions such as language model and lexical length model, positive phrase translation model and positive lexical weight model are all related in machine translation systems.

### 3.2 Coordinate systems

In the KCD approach, feature functions are directly used by the iterative search process. When searching for a better scoring point, all the search directions do not interfere with each other. However, most of the feature functions in machine translation system are interrelated. In fact, there is a default insufficient assumption that each search direction corresponds to one feature function.

As alternative to KCD, in RSS-MERT, one direction corresponding to one feature function is only a basic form of the coordinate system. Coordinate system (RS) is consisted of a number of search directions which can be moved from one feature function near to another. This improved method helps to break the original assumption of feature function independence, and enhance the links between context-sensitive feature functions in the search process to get better weight set that is closer to the real distribution of feature functions.

Figure 1 compares the difference between KCD and RSS-KCD.

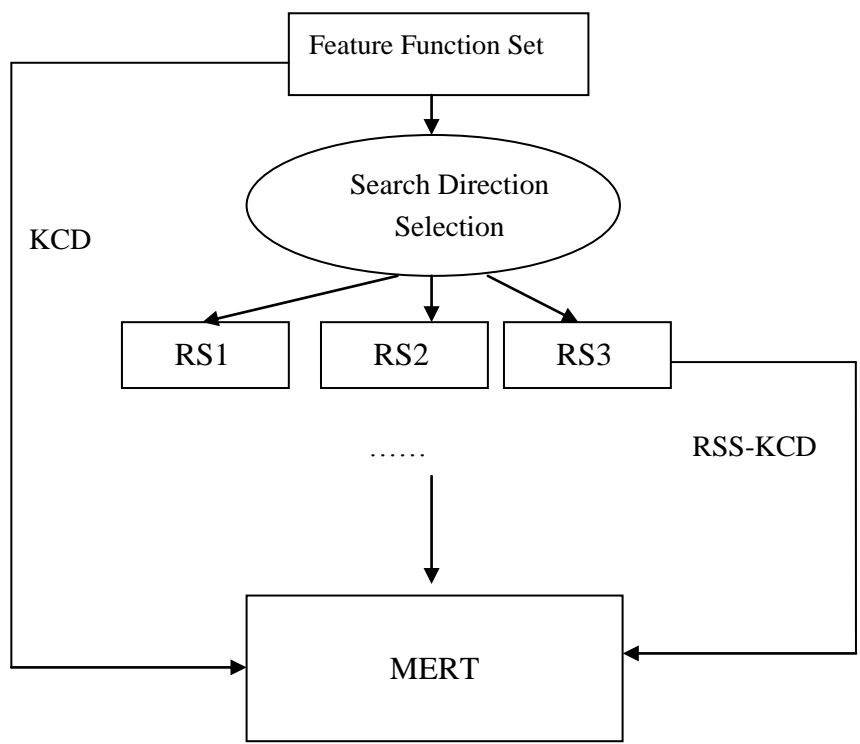

Figure 1    RSS-KCD adds a coordinate system to KCD

### 3.3 RSS-MERT Progress

RSS-MERT algorithm is generated by adding selected coordinate systems to KCD. The association between two functions will change according to some objective factors (such as training corpus). Therefore, in RSS-MERT process, we need the results of closed test to adjust the coordinate system. The premise that we select coordinate system according to the closed test results is the consistency between the closed and open test results. We change coordinate system by moving search directions of three context-sensitive feature functions. The experiments show that it is feasible to select coordinate systems by the closed test results. Algorithm 2 presents RSS-MERT progress in details.

Algorithm 2  Adding Coordinate systems to Koehn's Coordinate Descent Method

  1 Select two associated feature functions A and B;
  2 for $\alpha \leftarrow$ -1 to 1 step 0.1 (initial value, final value and step size can be adjusted according to the actual require)
  3 Initialize the coordinate system: let each dimension corresponds to one feature function
  4 rotate dimension A to dimension B by $arctg\alpha°$ to form a new coordinate system
  5 iter=1;lasterror=0;newerror=1;epsilon=0.001
  6 While (iter<Maxiter and abs(lasterror-newerror)>epsilon)
  7     for d=1 to M // for each dimension
  8         for s=1 to S // for each sentence in training corpus
  9             Compute critical value and delta error
 10         endfor
 11         merge all critical value and sort them
 12         compute error within each pair of non-identical boundaries
 13         select the weight $\lambda$ as the midpoint of the interval corresponding to the lowest error
 14     end for
 15     iter++
 16     lasterror=newerror
 17     newerror=lowest error
 18 end while
 19 do closed test and open test for the current model
 20 end for
 21 Select the best $\alpha$

## 3.4 Establishing Coordinate system

Coordinate system is established based on search directions. Each search direction can be moved from one feature function to another. In order to add function associating factors, we introduce a parameter $\alpha$ based on the original function $h(e|f_s)$, combining with another feature function $h'(e|f_s)$ to form a dynamic search direction $D(e|f_s)$:

$$D(e|f_s) \leftarrow h(e|f_s) + \alpha h'(e|f_s) \quad (4)$$

Dynamic search direction is determined by three parts: the main feature function $h(e|f_s)$, the subsidiary feature function $h'(e|f_s)$, and the parameter $\alpha$. $h(e|f_s)$ and $h'(e|f_s)$ are two interrelated feature functions, and the relationships between them are determined by the parameter $\alpha$.

In the original system, each search direction is in accord with one of the feature function. By introducing the parameters $\alpha$, the search direction can be moved to another feature, so as to create a new coordinate system (see Figure 3).

The establishment of dynamic search directions makes use of a parameter to consider the intrinsic link between feature functions. By changing the parameter, the machine translation system can construct several coordinate systems and use the error rate to adjust the feature weight proportion on different RS. The local optimum point will change while search direction is moved, thus we have the chance to obtain a weight set of translation models with higher evaluation results.

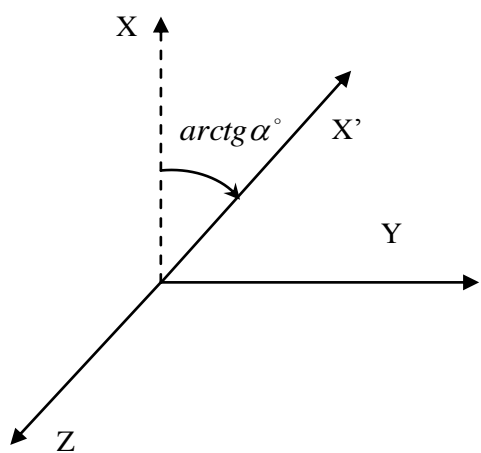

Figure 3 Assume that the original system has three feature functions, and coordinate system is XYZ. Through movement of one search direction X, the new coordinate system is changed to X'YZ.

## 4 Results

### 4.1 Baseline Translation System

We use a free and complete machine translation decoding tool Moses[2] as the baseline system. Moses is a phrase-based statistical machine translation system developed by Philipp Koehn et al. The whole system is written in C++ language. From training to decoding, the code is completely open source, and can be run on both Linux and Windows platforms.

Moses is no longer confined to the phrase-based translation table, but is a translation model that can integrate part of speech, word type, word stemming and other features. Moses uses a hybrid network decoding technology that allows a variety of possible input forms, such as the output of named entity or speech recognition.

In addition, when using Moses we must also install the language model tool SRILM[3] and word alignment tool GIZA++[4].

### 4.2 Corpus Preparation

Our data come from two kinds of parallel corpus. One is the corpus of aligned sentences in Chinese and English collected by Computing Technology Institute of the Chinese Academy of Science. It including 11 fields such as arts, economy, legal, environment, life, politics, science etc, which are all sentence-level aligned. Another is 863 Chinese-English sentence aligned corpus.

### 4.3 Training and Development Data

Chinese and English corpus used in the translation model has 86,933 sentences. Additionally, there are 187,014 translation dictionary entries. The English corpus used by language model is consisted by 251909 sentences.

The Chinese-English test corpus is provided by the Fourth National Machine Translation Conference (CWMT2008). It has a total of 1006 Chinese sentences. Each sentence corresponds to 4 English translations.

For parameter adjusting, we use Chinese-English translation corpus offered by NIST2008. It has a total of 691 Chinese sentences. Each Chinese sentence corresponds to 4 English translations.

---

[2] http://www.statmt.org/moses/

[3] ftp://ftp.speech.sri.com/pub/people/stolcke/srilm/srilm-1.5.7.tar.gz

[4] http://code.google.com/p/giza-pp/downloads/list

## 4.4 Results

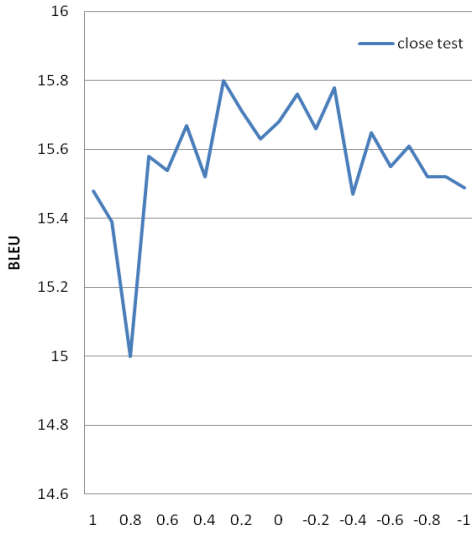

Figure 4  Closed test results (move from language model direction to distortion model direction，ngram=6, the horizontal coordinate denotes $\alpha$ )

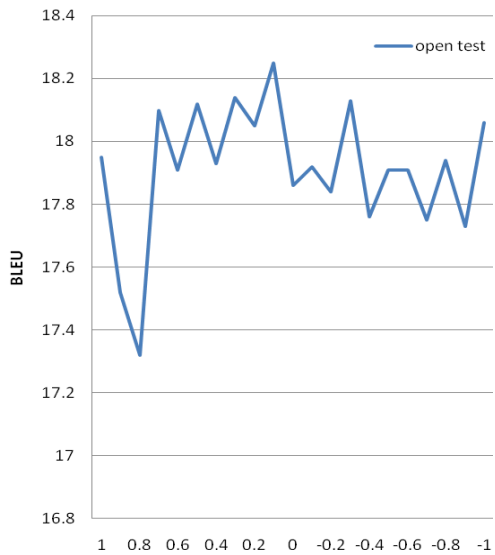

Figure 5   Open test results (move from language model direction to distortion model direction，ngram=6, the horizontal coordinate denotes $\alpha$ )

Figures 4 and 5 present the closed and open test results respectively when the search direction is moved from language model to distortion model and n-gram is 6. The results show that when the parameter is greater than 0, the open test and closed test results are significantly better than when the parameter is less than 0. Closed test and open test are approximately consistenct: closed test results reach a highest point when the parameter is 0.3, while the open test reach a highest point when the parameter is between 0.1 and 0.3. At the time the parameters is 0.3, the BLEU score of the closed test is 15.80, and of the open test is 18.14, significantly higher than the baseline of 15.68 and 17.86.

Table 1 and table 2 compare the best results of three different kinds of search direction movements with the baseline system when n-gram is 5 and 6 respectively. Three movements including (1) move from language model direction to distortion model direction (2) move from language model direction to length model direction (3) move from distortion model direction to length model direction. As the closed test and open test evaluation results have a moderate consistency, at the best result point in the closed test, open test result also improves significantly than the baseline system. Experimental results show that RSS-MERT method can effectively overcome the limitation that the iterative method always converges to a local optimum, and make the open test and closed test results improved significantly.

|  | baseline | lm to wp | dm to wp | lm to dm |
|---|---|---|---|---|
| $\alpha$ |  | -0.7 | +0.7 | +0.1 |
| open test | 17.75 | **17.99** | 17.92 | 17.98 |
| close test | 15.52 | 15.69 | 15.61 | **15.73** |

Table 1  Comparing the best results of three different kinds of movements with the baseline system when n-gram is 5

|  | baseline | lm to wp | dm to wp | lm to dm |
|---|---|---|---|---|
| $\alpha$ |  | +0.4 | -0.2 | +0.3 |
| open test | 17.86 | 17.90 | 18.10 | **18.14** |
| close test | 15.68 | 15.76 | **15.84** | 15.80 |

Table 2  Comparing the best results of three different kinds of movements with the baseline system when n-gram is 6

In addition, the two tables show that the experiment results of 6-gram is significantly better than that of 5-gram, which is consistent with the baseline system. When moving from language model direction to distortion model direction, the system achieved the best results: the BLEU score of closed test is 15.80 and of open test is 18.14.

We also try to move two search directions: one is from language model to distortion model and the other is from distortion model to length model. The parameters are 0.3 and 0.1 respectively. The results are showed below.

|  | open test | close test |
|---|---|---|
| MERT | 17.86 | 15.68 |
| 1 dimension RSS-MERT | 18.14 | 15.80 |
| 2 dimension RSS-MERT | **18.19** | 15.81 |

Table 3  the results of RSS-MERT

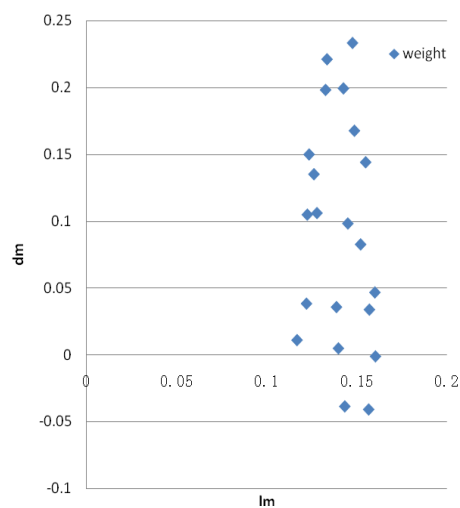

Figure 6 Two-dimensional map of changed language model and length model weight (move from language model direction to distortion model direction, ngram=5)

It can be seen from figure 6 that while transforming the search direction of the coordinate system, there is also a corresponding movement of the weights of two related features in the coordinate system. Therefore, changing the coordinate system is actually equivalent to moving the weight values in the coordinate system. As a whole, in RSS-MERT, by changing the search direction of the coordinate system and making the minimum error rate point moved, we can avoid the training progress converge to a local optimum.

## 5 Related Research

Minimum error training (MERT) is a

subject that still needs to be explored further. Cer (2008) proposes regularization and random search to ease the problem of local optimum. They select random search directions based on Powell method (Press, 2007) to overcome the limitation that the assumption it uses to build up the diagonal search directions do not hold in the present context . In addition, by checking and combining the adjacent peak in a fixed window of search directions, they solve the problem that one optimal value is surrounded by several bad objective function values in minimum error rate training. Their results show that these two improved methods can significantly improve the system performance compared to traditional Powell algorithm and coordinate descent method. Machery (2008) replaces word n-best list by lattices to improve the estimates of the expected translation score. In the algorithm, they efficiently construct and represent the exact error surface of all the translations which are encoded in a phrase lattice. They use this novel method to train the feature function weights for a phrase-based statistical machine translation system. Experiments show that they obtain significant improved BLEU score over N-best MERT.

The optimal solution point in MERT is often related to the initial weights of feature functions. Some current studies focus on the choice of initial weights. In order to avoid getting into a bad local optimum, Koehn et al. (2007) start from different initial weights, and use multiple optimizing search methods on each set of extended hypotheses in the translation. Moore et al.(2008) use two methods to adjust the initial weight for MERT: (1) randomly select initial weight according to uniform distribution; (2) use random walk method that select the initial weight according to the local optimum which the previous iteration converges to. Their experiments also make the BLEU evaluation results of translations improved significantly.

# 6 Conclusion

Statistical machine translation systems often contain a number of feature functions. So the problem of correlated feature functions is critical that any statistical machine translation system must face. However, in Koehn coordinate descent method, each search direction corresponds to one feature function, ignoring the correlation between the feature functions. In this paper we change the search directions to form several coordinate systems for KCD so that we can get better weight points which are closer to the real distribution of feature functions.

The weights of minimum error rate training tend to converge to a local optimum. Experiments show that the changing trends of feature function weights are the same as the moving trends of search directions. By making use of the correlation between the functions and selecting coordinate systems, we can obtain multiple weight sets that satisfy minimum error rate. For different weight sets, the closed test and open test results of BLEU evaluation are often identical. Therefore, we do not need to change the initial weights of training, instead we only need to select the best model for the closed test, so as to improve the open test results. This method overcomes the limitation of the local optimum in MERT and can get better results on the overall situation.

Without increasing the system's execute time and storage space, this method is simple and convenient to be widely used. Moreover, the SMT performance has a chance to be significantly improved by finding better coordinate systems. In the future, we will extend RSS-MERT methods to other models, such as phrase translation model and lexical weight model. In addition, we will also test on a larger corpus, hoping to expand such a simple way to the practical aspects.

## Acknowledgments

This work was supported by 2010 Jiangsu Graduate Research Innovation Plan CX10B_063R and 2010 Outstanding Ph.D. Dissertation Breeding Program in Nanjing Normal University 2010bs0006.